\documentclass[runningheads]{llncs}
\usepackage[T1]{fontenc}
\usepackage{graphicx}
\begin{document}
\title{Parameter-Efficient Fine-Tuning (PEFT) of Vision Foundation Models for Atypical Mitotic Figure Classification}

\titlerunning{PEFT for Atypical Mitotic Figure Classification}

%
\author{Lavish Ramchandani\inst{1} \and
Gunjan Deotale\inst{1} \and Dev Kumar Das\inst{1}}
\authorrunning{Ramchandani et al.}
%
\institute{Aira Matrix Private Limited, Mumbai, India, 
\email{info@airamatrix.com}\\
\url{https://www.airamatrix.com/}}
\maketitle              
\begin{abstract}

Atypical mitotic figures (AMFs) are rare abnormal cell divisions associated with tumor aggressiveness and poor prognosis. Their detection remains a significant challenge due to subtle morphological cues, class imbalance, and inter-observer variability among pathologists. The MIDOG 2025 challenge introduced a dedicated track for atypical mitosis classification, enabling systematic evaluation of deep learning methods. In this study, we investigated the use of large vision foundation models, including Virchow, Virchow2, and UNI, with Low-Rank Adaptation (LoRA) for parameter-efficient fine-tuning. We conducted extensive experiments with different LoRA ranks, as well as random and group-based data splits, to analyze robustness under varied conditions. Our best approach, Virchow with LoRA rank 8 and ensemble of three-fold cross-validation, achieved a balanced accuracy of 88.44\% on the test set, ranking 9th in the challenge. These results highlight the promise of foundation models with efficient adaptation strategies for the classification of atypical mitosis, while underscoring the need for improvements in specificity and domain generalization.

\keywords{MIDOG 2025  \and Atypical Mitoses \and Foundation Models \and  Classification \and Low-Rank Adaptation}
\end{abstract}
\section{Introduction}
Mitotic figures (MFs) are cells undergoing division and are widely used as histopathological biomarkers of tumor proliferation and grading. Traditional approaches have primarily relied on counting the number of mitotic figures per unit area, which has been integrated into standard grading systems such as the Nottingham histological grading for breast carcinoma. However, recent studies emphasize that not only the quantity of mitotic figures but also the presence of atypical mitotic figures (AMFs) characterized by abnormal morphologies such as multipolar spindles, chromatin bridges, or lagging chromosomes may provide additional prognostic value in breast cancer and canine tumors.

The identification of AMFs is notoriously difficult. Their frequency is low compared to normal mitotic figures (NMFs), and the morphological cues can be subtle and subjective. Inter-observer variability among pathologists remains high, making manual annotation time-consuming and inconsistent. Automated computational methods are therefore essential to scale annotation and reduce subjectivity.

The MIDOG 2025 \cite{Ammeling2025} challenge (Track 2: Atypical Classification) provides a benchmark dataset designed to evaluate algorithms for differentiating AMFs from NMFs across multiple domains. This task is challenging due to severe class imbalance (far fewer atypical figures than normal ones), heterogeneous domains (human vs. canine tumors, different scanners and labs), and the inherently subtle distinction between classes.

Deep learning has recently shown promise in this area. In particular, vision foundation models such as Virchow \cite{vorontsov2024virchowmillionslidedigitalpathology} (a DINOv2-pretrained \cite{oquab2024dinov2learningrobustvisual} ViT-H/14 model on histology slides) have emerged as strong backbones for pathology tasks. Prior work \cite{banerjee2025benchmarkingdeeplearningvision} demonstrated that parameter-efficient fine-tuning approaches such as Low-Rank Adaptation (LoRA) \cite{hu2021loralowrankadaptationlarge}  can substantially improve adaptation of large models to specific tasks.

Motivated by these advances, we explored UNI \cite{chen2023generalpurposeselfsupervisedmodelcomputational}, Virchow and Virchow2 \cite{zimmermann2024virchow2scalingselfsupervisedmixed} foundation models, trained with LoRA-based fine-tuning and evaluated using both random and domain-aware splits. Our contributions are summarized as follows:
\begin{enumerate}
    \item Benchmarking  UNI, Virchow and Virchow2 with different hyperparameters.
    \item Analysis of split strategies to study domain generalization.
    \item Submission of our best approach (Virchow + LoRA rank 8) to MIDOG 2025, achieving 88.44\% balanced accuracy (9\textsuperscript{th} place) on the test set.
\end{enumerate}

\section{Methods}

\subsection{Dataset}

We used the official dataset from MIDOG 2025 Track 2: Atypical Classification, which integrates data from AMi-Br \cite{palm_histologic_2025}, MIDOG 2025 \cite{weiss_2025_15188326}, and OMG-Octo \cite{Shen2025} Atypical datasets. These datasets consist of mitotic figure crops (AMFs or NMFs) annotated by pathologists from hematoxylin and eosin (H\&E) stained slides. The MIDOG 2025 dataset provides 10,191 normal mitotic figures and 1,748 atypical mitotic
figure annotations across 454 labeled images from 9 distinct domains, making it one of the most diverse datasets for mitotic figure analysis.

\subsection{Models}
We investigated three foundation models:

\begin{itemize}

\item UNI: A ViT-L/16 model pre-trained across multimodal pathology datasets.

\item Virchow: Vision Transformer (ViT-H/14) pre-trained on > 1M Whole Slide Images (WSIs) using DINOv2.

\item Virchow2: Vision Transformer (ViT-H/14, 632 M parameters) pre-trained with DINOv2 on 3.1 million histopathology WSIs.
\end{itemize}
These models were adapted to binary classification (AMF vs NMF) via a linear classifier head.


\setlength{\tabcolsep}{5pt}

\begin{table}[t]
\caption{Performance of foundation models across different LoRA ranks, splits, and upsampling strategies.}
\resizebox{\textwidth}{!}{
\begin{tabular}{lcllcp{3cm}p{3cm}}
\hline
Foundation Model     & LoRA Rank    &      Split    & Upsampling & Threshold &
\begin{tabular}[c]{@{}c@{}}Average Validation \\ Balanced Accuracy\end{tabular} &
\begin{tabular}[c]{@{}c@{}}Balanced Accuracy\\ (Preliminary Test Set)\end{tabular} \\
\hline
UNI      & 8 & Group Split  & Padding to 224 & 0.5 & 0.8037 & 0.7847 \\
Virchow  & 8 & Group Split  & Padding to 224 & 0.5 & 0.8559 & 0.8387 \\
Virchow2 & 8 & Group Split  & Padding to 224 & 0.5 & 0.8091 & -      \\
Virchow  & 4 & Group Split  & Padding to 224 & 0.5 & 0.8587 & -      \\
Virchow  & 8 & Random Split & Padding to 224 & 0.5 & 0.8673 & 0.844  \\
Virchow  & 8 & Random Split & Resized to 224 & 0.5 & \textbf{0.8682} & 0.8612 \\
Virchow  & 8 & Random Split & Resized to 224 & 0.6 & 0.857  & \textbf{0.8837} \\
\hline
\end{tabular}
}
\label{tab:foundation}
\end{table}

\subsection{Fine-Tuning with LoRA}
Fine-tuning large ViTs on limited pathology data risks overfitting. We therefore applied Low-Rank Adaptation (LoRA), which injects trainable rank-decomposition matrices into transformer attention layers. LoRA enables efficient adaptation with fewer parameters while preserving pretrained weights.

We tested ranks 4 and 8 focusing on attention projections (query/key/value). Dropout of 0.3 and scaling factor 16 were applied. Rank 8 provided the best trade-off between expressivity and regularization.

\subsection{Cross-Validation and Splits}
We performed three-fold cross-validation to ensure robust estimation:
\begin{itemize}
\item Random split: Images randomly assigned across folds.

\item Group/domain split: All images from a given tumor type are assigned entirely to either training or validation set.

\end{itemize}
This allowed us to test in-domain vs cross-domain generalization.

\subsection{Training Setup}
\begin{itemize}
\item Input resolution: Two strategies were evaluated 
(i) padding patches to 224×224, and (ii) resizing patches to 224×224, since all foundation models were pretrained on inputs of size 224×224.

\item Augmentations: random flip, rotation, color jitter, and resized crop.

\item Optimizer: AdamW \cite{kingma2017adammethodstochasticoptimization} (lr=5e-4, weight decay=1e-2).

\item Binary cross-entropy with logits loss (BCEWithLogitsLoss).

\item Sampling strategy: To address class imbalance, a Weighted Random Sampler was used to ensure approximately balanced batches during training.

\item Scheduler: ReduceLROnPlateau (factor=0.5, patience=3).

\item Early stopping: patience=10 epochs (monitoring validation Balanced Accuracy).

\item Batch size: 8.
\end{itemize}

\subsection{Evaluation and Inference Strategy}
As per the challenge, we report Balanced Accuracy (BAC) as the primary metric, alongside Sensitivity, Specificity, and AUROC. 

By default, binary classifiers use a threshold of 0.5 to assign class labels. However, this value is not necessarily optimal for our target metric, balanced accuracy (BAC), especially in imbalanced or domain-shifted datasets. During validation, we therefore evaluated multiple thresholds between 0.35 and 0.75 and selected the one that maximized BAC. Preliminary leaderboard feedback showed that our models achieved very high sensitivity but comparatively low specificity when using 0.5. To address this, we adopted a stricter threshold of 0.6 for the positive (atypical) class, which reduced false positives and led to a higher BAC. We employed an ensemble of models trained on three different folds on the test set.


\setlength{\tabcolsep}{5pt}  

\begin{table}
\caption{Domain-wise ROC AUC, Accuracy, Sensitivity, Specificity, and Balanced Accuracy on the preliminary test set.}
\label{tab:domain_performance}
\begin{tabular}{llllll}
\hline
Domain & ROC AUC & Accuracy & Sensitivity & Specificity & Balanced Accuracy \\
\hline
Domain 0 & 0.8906 & 0.8055 & 0.7500 & 0.8125 & 0.7812 \\
Domain 1 & 0.9273 & 0.8385 & 0.9310 & 0.8181 & 0.8746 \\
Domain 2 & 0.9584 & 0.8400 & 1.0000 & 0.7752 & 0.8876 \\
Domain 3 & 1.0000 & 0.8684 & 1.0000 & 0.8611 & 0.9305 \\
Overall  & 0.9484 & 0.8388 & 0.9577 & 0.8096 & 0.8837 \\
\hline
\end{tabular}
\end{table}

\section{Results}

Table \ref{tab:foundation} summarizes the performance of UNI, Virchow and Virchow2 across different LoRA ranks, thresholds, upsampling and split strategies. 

On the preliminary test set, our best submission (Virchow with LoRA rank 8, three-fold cross-validation) achieved a balanced accuracy of 88.37 \%, placing joint 9th in the preliminary leaderboard and subsequently achieved 88.44 \% balanced accuracy ranking 9th on the final test set.

The preliminary test set consisted of data from 4 different domains. Table \ref{tab:domain_performance} presents the detailed performance metrics across
different domains and overall performance of our best submission.

\section{Discussion}

Our results confirm the effectiveness of Virchow with LoRA for atypical mitosis classification. Compared to random splits, group/domain splits exposed greater variability, underscoring the challenge of domain shift. Virchow consistently achieved strong sensitivity, aligning with prior benchmarks \cite{banerjee2025benchmarkingdeeplearningvision}, but specificity remained lower, suggesting difficulty in correctly rejecting normal mitoses.

Exploring UNI and Virchow2 highlighted that while alternative foundation models provide reasonable performance, Virchow remains competitive and better aligned with the challenge dataset resolution and domain characteristics.

\section{Conclusion}

We presented a foundation model-based approach for atypical mitosis classification in the MIDOG 2025 challenge. Using Virchow with LoRA rank 8 and three-fold cross-validation, we achieved 88.44\% balanced accuracy on the test set, ranking 9th. Our study highlights the importance of parameter-efficient adaptation, careful cross-validation, and evaluation under domain shifts.


\begin{credits}
\subsubsection{\ackname} We thank the MIDOG 2025 challenge organizers for providing this valuable dataset and evaluation platform, which
advances the field of computational pathology and automated mitotic figure analysis.
\end{credits}

\bibliographystyle{splncs04}
\bibliography{ref2.bib}
%





\end{document}